\documentclass[10pt,twocolumn,letterpaper]{article}

\usepackage{cvpr}
\usepackage{times}
\usepackage{epsfig}
\usepackage{graphicx}
\usepackage{amsmath}
\usepackage{amssymb}

\usepackage[breaklinks=true,bookmarks=false]{hyperref}

\cvprfinalcopy % *** Uncomment this line for the final submission

\def\cvprPaperID{****} % *** Enter the CVPR Paper ID here

\begin{document}

%%%%%%%%% TITLE
\title{Iterative Self-Learning: Semi-Supervised Improvement to Dataset Volumes and Model Accuracy}

\author{Robert~Dupre, Jiri~Fajtl, Vasileios~Argyriou, Paolo~Remagnino\\
Kingston University, Penrhyn Rd\\
{\tt\small }
}
\maketitle
%\thispagestyle{empty}

%%%%%%%%% ABSTRACT
\begin{abstract}
A novel semi-supervised learning technique is introduced based on a simple iterative learning cycle together with learned thresholding techniques and an ensemble decision support system. State-of-the-art model performance and increased training data volume are demonstrated, through the use of unlabelled data when training deeply learned classification models. Evaluation of the proposed approach is performed on commonly used datasets when evaluating semi-supervised learning techniques as well as a number of more challenging image classification datasets (CIFAR-100 and a 200 class subset of ImageNet).

\end{abstract}
\vspace{-8mm}

%%%%%%%%% BODY TEXT
\section{Introduction}

Semi-supervised learning has become one of the most prevalent topics within image processing and computer vision research in recent years. With the ever increasing availability of high powered GPU hardware and the success of deep learning on applications such as computer vision \cite{he2016, Krizhevsky2012}, speech recognition \cite{Graves2014}, and natural language processing \cite{sutskever2014, goldberg2017} the need for large scale datasets to support these methods becomes a higher priority, as well as a bottleneck to performance improvement. Semi-supervised processes have been applied successfully in many areas such as image classification and segmentation \cite{Hong2015}, natural language processing and artificial intelligence \cite{Carlson2010}. Typically, semi-supervised deep learning uses novel model architectures, regularization methods or loss functions combining outputs from known labels with unknown labels to provide more accurate outputs, \cite{haeusser2017learning, cicek2018saas}. Laine \cite{Laine2017} utilises an architecture based on ensemble predictions, acquired during the training of a network at different epochs or under different regularization and input conditions. Tarvainen et al. \cite{Tarvainen2017} take the concept of temporal ensembling and extend it to the model weights. French et al. \cite{french2018self} extend by introducing class balancing and confidence thresholding. Miyato et al. \cite{Miyato2017} consider a novel regularization method to semi-supervised learning.

%In much of the current work in this area the training methods, architectures and loss functions utilised have been %developed specifically for that semi-supervised application, making them specific to task and almost impossible to %transfer to a new area.
The following method is an extension of \cite{rob2019tip}. The method iteratively reclassifies a dataset such that the model being trained is only ever exposed to what it considers fully labelled data. Firstly, a simple and easily implemented semi-supervised learning framework, independent from model architecture or loss functions making it applicable to a wide range of classification tasks. Secondly, novel learned thresholding techniques and metrics to supervise the dataset growth, ensuring only confidently classed samples are added to a training dataset.

\section{Methodology}\label{sec:Method}
The core assumption in this work is that generalization error always decreases with more training samples as shown by  \cite{barron1994approximation} and recently \cite{bartlett2017spectrally}.
To address this problem the Iterative Learning-Ensemble (IL-E) approach is presented. The iterative nature of the IL-E is given by the train, classify, analyse and finally update cycle. Firstly, a model ($\mathbf{\theta}$) is trained on a cleanly labelled dataset, $\mathbb{D}_l$, and validated on the $\mathbb{D}_v$ dataset. The training of the model is performed in a relevant way to the application and task, neither the architecture nor the loss functions are changed in any way. Secondly, the unlabelled samples are classified and the process of updating the training set is run.

Let $x \in \mathbf{R}^{d}$ represent an input variable in $d$ dimensions and $y \in \mathbf{L}^C$ represent the label associated with that sample, where $C$ represents the number of possible class labels. In this work, $x_i$ represents an image and $y_i^c$ the label from $C$-classes. From the pool of cleanly labeled and unlabelled data, three datasets are constructed: Labelled ($\mathbb{D}_l = {x_n^l, y_n^l | n = 1, \ldots, N^l}$), derived from a portion of the cleanly labelled data. Unlabelled ($\mathbb{D}_{u} = {x_m^{u}, y_m^{u} | m = 1, \ldots, M^{u}}$), indexed from only unlabelled data and validation ($\mathbb{D}_v = {x_o^v, y_o^v | o = 1, \ldots, O^v}$), derived from the remaining subset of the cleanly labelled data.

The primary issue when adding newly labelled samples to the training dataset is ensuring the model is confident that the additions are labelled correctly. This confidence is achieved in two ways: firstly, well established ensembling techniques are utilised to produce better predictions from a trained model \cite{Laine2017, Rasmus2015, Tarvainen2017} and secondly, a novel set of confidence metrics have been devised based solely on the posterior probabilities produced from model $\mathbf{\theta}$. Importantly, there are no additional clustering or preprocessing steps applied to the unlabelled data of any kind, the only assumption made within this work is that the data is of a similar quality, context and application as that within the cleanly labelled.

The goal of ensembling is to find the most positive class distribution for use with the confidence metrics. To this end, a number of augmentations are applied to an unlabelled sample such that $\mathbb{D}_{aug} = {\mathbf{x}^{(j)}_{a}, y^{(j)} | a = 1, \ldots, A^j}$ represents a single sample $\mathbf{x}^{(j)}$, augmented in $A$ different ways, each with the same label $y^{(j)}$. The augmented samples, including the original, are now passed to the model for inference and the posterior probability vectors, or class distributions, for all the augmented samples $\tilde{z}$ returned.

\begin{equation} \label{eq:12}
    \tilde{z} = P({y}|\mathbb{D}_{aug}; \theta)
\end{equation}

The returned posterior probability vectors are then scaled by the similarity of the class distributions returned as a result of Eq. \ref{eq:12}. The standard deviation of the posterior probabilities between class labels across the augmented samples is then subtracted from $\tilde{z}$ as form of scaling. Augmented ensembles which when evaluated differ greatly in their class distributions, result in a larger standard deviation. In turn, this would penalise the final confidence score $\mathbf{x}^{(j)}_{a}$ more so than when the model produces similar distributions across the ensemble. Finally, the augmented sample $a$ with the highest posterior probability for any class label, is selected and its original unscaled class distribution used as input $\mathbf{x}^{(j)}$ for the confidence metrics highlighted in Eq. \ref{eq:3}, \ref{eq:4} and \ref{eq:6}.

\vspace{-3mm}
\begin{equation} \label{eq:13}
    a = argmax_{a} (\tilde{z}_a - \sigma)
\end{equation}

The confidence metrics used to further ensure unlabelled samples are correctly labelled, cover three distinct areas computed from the posterior probabilities after evaluation of the unlabeled data $\mathbb{D}_u$, (see supplementary material for visual representations of these concepts). Firstly, the single highest class activation obtained from the posterior distribution $c_{a}$ (higher is better). Formally, consider an unlabeled sample $\mathbf{x^{(j)}} \in \mathbb{D}_u$

\begin{equation} \label{eq:2}
    {y_1},{y_2} = argmax_{{y}} P(\mathbf{y} | \mathbf{x^{(j)}};\theta)
\end{equation}

Where ${y_1}$ and ${y_2}$ are labels corresponding to the first and second highest posterior probabilities. The $c_{a}$ is then

\begin{equation} \label{eq:3}
    c_{a} = P({y_1} | \mathbf{x^{(j)}};\theta)
\end{equation}

\begin{figure*}[!ht]
	\begin{center}
        \includegraphics[width=.40\linewidth]{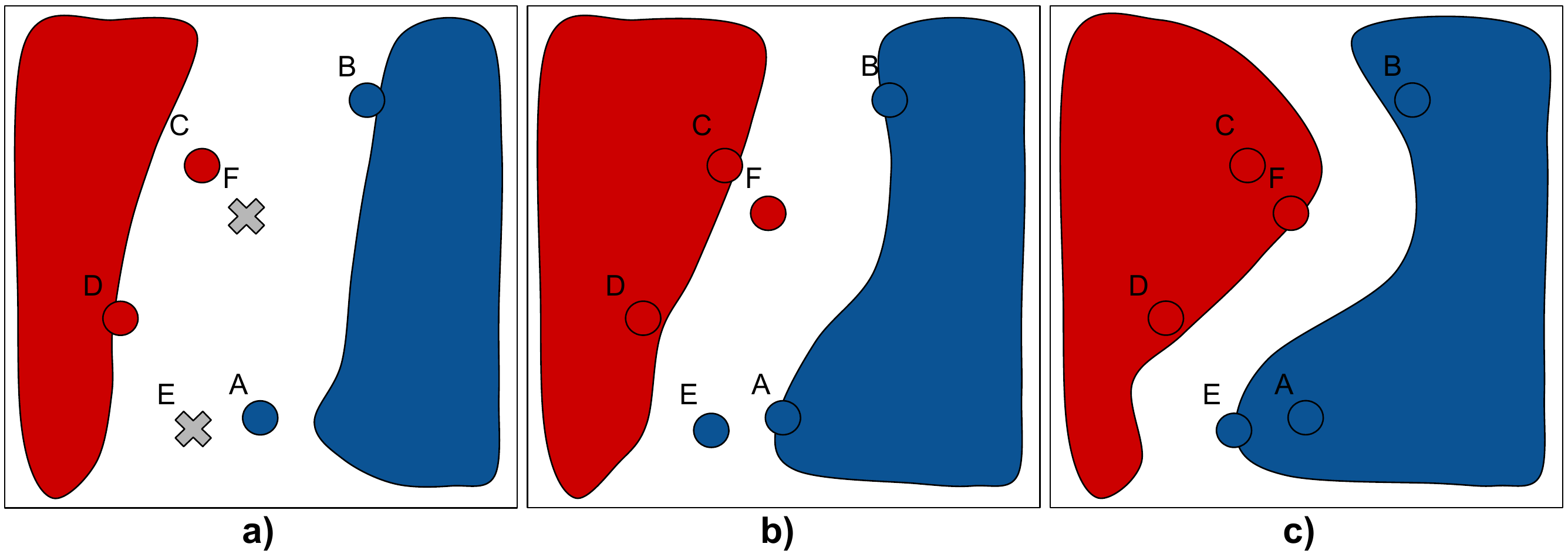}
        \includegraphics[width=.13\linewidth]{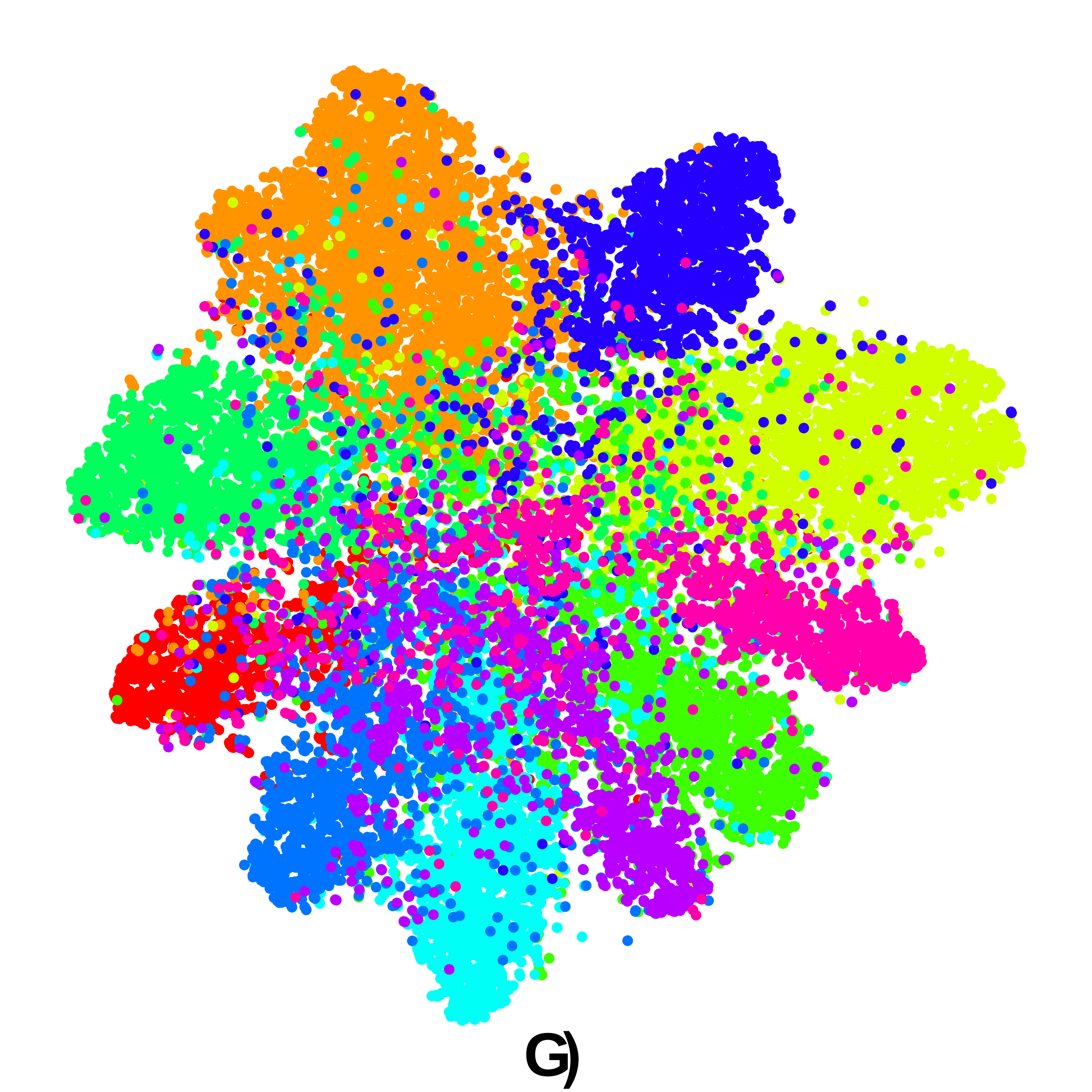}
        \includegraphics[width=.13\linewidth]{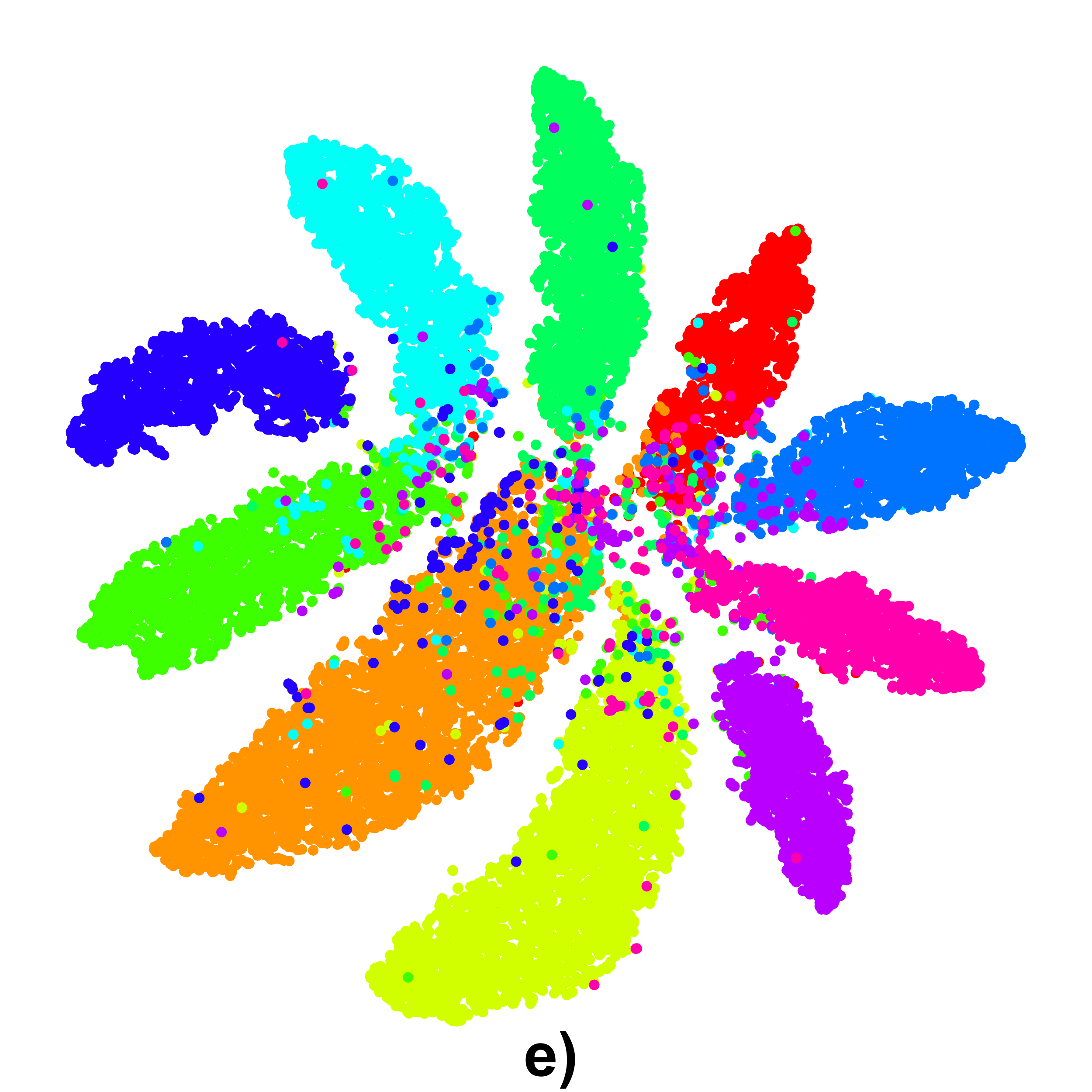}
	\end{center}
	\caption{(Left) A-F  $ \in \mathbb{D}_u$,   red and blue areas are two class manifolds  $ \in \mathbf{\theta}$, a) Iter 1:  A,B,C and D are classified as  red or blue using proximity to the respective manifolds, while E and F remain unclassified (confidence is not high enough). b) Iter 2: after retraining the model with new samples A-D, confident becomes sufficient to classify  E and F and add them to the new training set. c) Iter 3:  the manifolds are updated with the samples E and F. (Right) t-SNE  of the last fully connected layer (1024 neurons) of JFNet2  when evaluating the SVHN validation dataset. d) Clusters with the model trained on the initial 1000 samples. e) Clusters after IL-E has been run for 75 iterations, increasing the training dataset size and improving classification accuracy.}
	\label{fig:manifold_update}
%\vspace{-5mm}
\end{figure*}
Second, the difference between the highest and second highest activation $c_{b}$ (larger difference is better) is computed according to Eq. \ref{eq:4}.
\begin{equation} \label{eq:4}
    c_{b} = P({y_1} | \mathbf{x^{(j)}};\theta)-P({y_2} | \mathbf{x^{(j)}};\theta)
\end{equation}
Lastly, $c_{c}$ is calculated as the Euclidean distance between the posterior distribution for the unlabeled sample $\mathbf{x}^{(j)}$ and the average distribution $p_t(y_1)$ for the predicted class $y_1$ (lower score is better). $p_t(y_1)$ is computed over all training samples of class $y_1$. These average distributions per class are computed at the end of each model training iteration and are recorded for use in these confidence computations.
\begin{equation} \label{eq:6}
    c_{c} = \lVert P(\mathbf{y} | \mathbf{x}^{(j)};\theta) - p_t(y_1) \rVert
\end{equation}
For each of these three metrics a value is returned, in the cases of $c_a$ and $c_b$ the value returned by the model should be high and for $c_c$ the distance between the two posterior probability distributions should be low, however the $c_c$ scores are inverted so as to have a uniform, higher is better policy. The weighted sum of these metrics scores is then used to provide a final confidence score for a specific unlabeled sample $\mathbf{x}^{(j)}$. As some metrics are more informative that others their contribution to the final confidence $c$ should reflects this. The weighting is found experimentally but is rooted on the accuracy of the metric on a set of unlabelled samples. Importantly these values may change based on application as certain metrics may be more informative in different problems.
\begin{table*}
    \centering
    \caption{IL-E results across the three datasets (SVHN, CIFAR-100, TinyImageNet),  with full and subset benchmark training results.}
    \label{tab:Results_float}
    \scalebox{0.65}{
\begin{tabular}{ccccc}
                                                & Error Rate\% ($\sigma$)   & Error Rate \% (Improvement)   & Error Rate\% ($\sigma$) & Added Samples (Acc. \%) \\ \cline{2-5}
                                                &                           & SVHN                          &                         &                             \\
Model                                           & 1k Benchmark              & 1k Samples                    & Full Benchmark          &                             \\ \hline
\multicolumn{1}{c|}{GAN \cite{Salimans2016a}}   & N/A                       & 8.11\%                        & N/A                &  -                           \\
\multicolumn{1}{c|}{$\prod$ model \cite{Laine2017}}& N/A                    & 4.82\%                        & 2.54\% ($\pm$0.04) &  -                           \\
\multicolumn{1}{c|}{Temporal E. \cite{Laine2017}}& N/A              & 4.42\%                        & 2.74\% ($\pm$0.06) &  -                           \\
\multicolumn{1}{c|}{VAT+EntMin \cite{Miyato2017}}& N/A                      & 3.86\%                        & N/A                &  -                          \\
\multicolumn{1}{c|}{ResNet-18 (IL-E)}           & 19.74\% ($\pm$0.32)       & \textbf{4.29\% (-15.45)}      & 2.98\% ($\pm$0.04) & 71,068 (94.89\%)            \\
\multicolumn{1}{c|}{LeNet-5 (IL-E)}             & 25.24\% ($\pm$1.55)       & 11.11\% (-14.13)              & 7.16\% ($\pm$0.09) & 42,999 (96.86\%)            \\
\multicolumn{1}{c|}{JFNet (IL-E)}               & 20.18\% ($\pm$0.50)       & 5.64\% (-14.54)               & 3.84\% ($\pm$0.05) & 66,421 (96.13\%)            \\ \hline
                                                &                           & CIFAR-100                     &                    &                             \\
                                                & 5k Benchmark              & 5k Samples                    & Full Benchmark     &                             \\ \hline
\multicolumn{1}{c|}{Temporal E. \cite{Laine2017}}& N/A              & 38.65\% (10k Samples)         & N/A                &  -                          \\
\multicolumn{1}{c|}{ResNet-18 (IL-E)}           & 32.49\% ($\pm$0.45)       & \textbf{28.09\% (-4.4)}       & 17.53\% ($\pm$0.09)& 42,526 (75.1\%)             \\
\multicolumn{1}{c|}{LeNet-5 (IL-E)}             & 89.21\% ($\pm$0.22)       & 87.47\% (-1.74)               & 65.55\% ($\pm$0.38)& 375 (72.53\%)               \\
\multicolumn{1}{c|}{JFNet (IL-E)}               & 39.66\% ($\pm$0.22)       & 66.49\% (-1.36)               & 39.66\% ($\pm$0.22)& 4,786 (73.21\%)             \\ \hline
                                                &                           & Tiny ImageNet                 &                    &                             \\
                                                & 10k Benchmark             & 10k Samples                   & Full Benchmark     &                             \\ \hline
\multicolumn{1}{c|}{ResNet-18 (IL-E)}           & 37.47\% ($\pm$0.46)       & \textbf{33.68\% (-3.79)}      & 27.38\% ($\pm$0.15)& 56,619 (81.37\%)            \\
\multicolumn{1}{c|}{LeNet-5 (IL-E)}             & 95.48\% ($\pm$0.43)       & 94.43\% (-1.05)               & 81.58\% ($\pm$0.27)& 69 (43.49\%)                \\
\multicolumn{1}{c|}{JFNet (IL-E)}               & 83.40\% ($\pm$0.12)       & 81.61\% (-1.79)               & 60.98\% ($\pm$0.25)& 684 (83.19\%)
\end{tabular}
}
%\vspace{-5mm}
\end{table*}
\begin{equation} \label{eq:7}
    c=c_{a}w_a+c_{b}w_b+\frac{1}{c_{c}}w_c
\end{equation}
Using a defined threshold $T_c$, samples can now be approved for inclusion in the labeled dataset $\mathbb{D}_l$, updated for use in the next training iteration. The threshold $T_c$ could be defined manually, allowing for policies where only very confidently analyzed samples are added or, through the use of a lower threshold, a more ``quantity over quality'' policy can be adopted. In this work the threshold value $T_c$ is learned. A process is run to find a threshold $T_c$, which when applied would add samples to $\mathbb{D}_l$ with a defined accuracy $T_a$, i.e defining a threshold $T_c$ whereby 99\% of samples added to $\mathbb{D}_l$ are correctly labelled. The process is run using only cleanly labelled data. The function $acc()$ calculates the percentage of correctly labeled samples $q$ in a dataset $\mathbf{X_u}$ classified by model $\mathbf{\theta}$ against their ground truth labels $\mathbf{y}$, given the threshold $T_c$.
\begin{equation} \label{eq:10}
    q = acc(\mathbf{X_u}, \mathbf{y} , T_c , \theta)
\end{equation}

Therefore given the required addition accuracy $T_a$ the max $T_c$ can be calculated,
%
%\vspace{-3mm}
\begin{equation} \label{eq:11}
    T_c = \max_{} {t_c}   \quad \text{subject to} \quad acc(\mathbf{X} , \mathbf{y} , t_c, \theta) > T_a
\end{equation}

Accuracy $T_a$ was set to $>99\%$. This process is run once on training data, as the model will be most confident on samples it has already seen and, as a result of this, impose a higher threshold than one defined using the validation set.
During these incremental updates the model is trained using an ever growing dataset. The dataset volume increases by the addition of unlabelled samples which the model has confidently identified belong to a respective class (i.e $> T_c$). As a result the model develops its \emph{knowledge} of specific classes and is therefore better able to identify additional samples in latter iterations. This process is symbolically shown in Figure \ref{fig:manifold_update} a-c, whereby a subset of new, unlabeled, samples get projected closer to the existing manifolds due to already learned characteristics of respective classes. Figure \ref{fig:manifold_update} d-e shows a real world example of the effect IL-E has had on the JFNet2 model's class manifolds. Additionally as the model is re-initialized at the beginning of each iteration, this method can leverage randomly initialised weights to help with the classification of unlabelled samples.
\vspace{-4mm}
\section{Results and Conclusions}\label{sec:Results}
% \subsection{Experiment Environment}
SVHN \cite{Netzer2011} is used for benchmarking and to better validate the performance of this iterative approach on a more challenging task,  CIFAR-100  \cite{Krizhevsky2009}, and a 200-class subset of ImageNet known as Tiny ImageNet are used. Initially benchmarks are run for each of the three models on the three datasets.  Table \ref{tab:Results_float} (columns 1 \& 3) outlines the benchmark error rates for each of these model architectures on both a subset of the training data and the full. The training subset size is based on 50 samples per class, CIFAR-100 uses 5,000 samples and Tiny ImageNet uses 10,000 samples. As the SVHN dataset is one of the most commonly used datasets when comparing semi-supervised learning techniques, the standard 1,000 samples is used (100 samples from each of the 10 classes). Each training subset is made up of an even distribution of classes with images from each class chosen at random. Each experiment was conducted four times with the average results presented along with the standard deviation given in brackets. The inclusion of these benchmarks is vital, especially for any result that uses a customised loss function or architecture, as without it is difficult to ascertain if improvement gains can be attributed to the model architecture used or the semi-supervised method.
%
% \section{Conclusion}
As demonstrated the simple iterative approach to semi-supervised learning IL-E has a number of benefits. Most notable being state of the art error rates on the CIFAR-100 dataset and near state of the art on SVHN dataset, achieved with no changes to the training methods, loss functions or model architectures used. The IL-E demonstrates, through the application of novel confidence metrics, the ability for a model to leverage its own confidence scores to improve classification accuracy.

\section{ACKNOWLEDGMENT}
This work is co-funded by the NATO within the WITNESS
project under grant agreement number G5437. The Titan X
Pascal used for this research was donated by NVIDIA.
%
% \pagebreak
{\small
\bibliographystyle{ieee}
\bibliography{SELF_LEARNING}
}
\end{document}